\documentclass[10pt,twocolumn,letterpaper]{article}

\usepackage{cvpr}              
\definecolor{cvprblue}{rgb}{0.21,0.49,0.74}
\usepackage[pagebackref,breaklinks,colorlinks,allcolors=cvprblue]{hyperref}
\usepackage{graphicx}
\usepackage{makecell}
\usepackage{bm}
\usepackage{booktabs}
\usepackage[table]{xcolor} 
\definecolor{headergray}{RGB}{235,235,235}
\usepackage{float}
\usepackage{pifont}
\usepackage{xurl}
\usepackage[para]{footmisc}
\usepackage[accsupp]{axessibility}  
\usepackage[a-3b]{pdfx}  

\def\paperID{9709} 
\def\confName{CVPR}
\def\confYear{2026}

\title{
HG-Lane: High-Fidelity Generation of Lane Scenes under \\ Adverse Weather and Lighting Conditions without Re-annotation
}

\author{
Daichao Zhao$^{1}$\footnotemark[1], 
Qiupu Chen$^{2}$\footnotemark[1], 
Feng He$^{3}$, 
Xin Ning$^{4}$, 
Qiankun Li$^{5}$\footnotemark[2] \\
$^{1}$Shanghai Jiao Tong University, $^{2}$School of Artificial Intelligence, Henan University, 
\\
$^{3}$University of Science and Technology of China, 
\\$^{4}$AnnLab, Institute of Semiconductors, Chinese Academy of Sciences,\\
$^{5}$Nanyang Technological University
}

\begin{document}
\twocolumn[{%
\renewcommand\twocolumn[1][]{#1}%
\vspace*{-3em}
\maketitle
\vspace{-1em}
\begin{center}
\setlength{\tabcolsep}{4pt}
\begin{tabular}{ccc}
\includegraphics[width=0.32\linewidth]{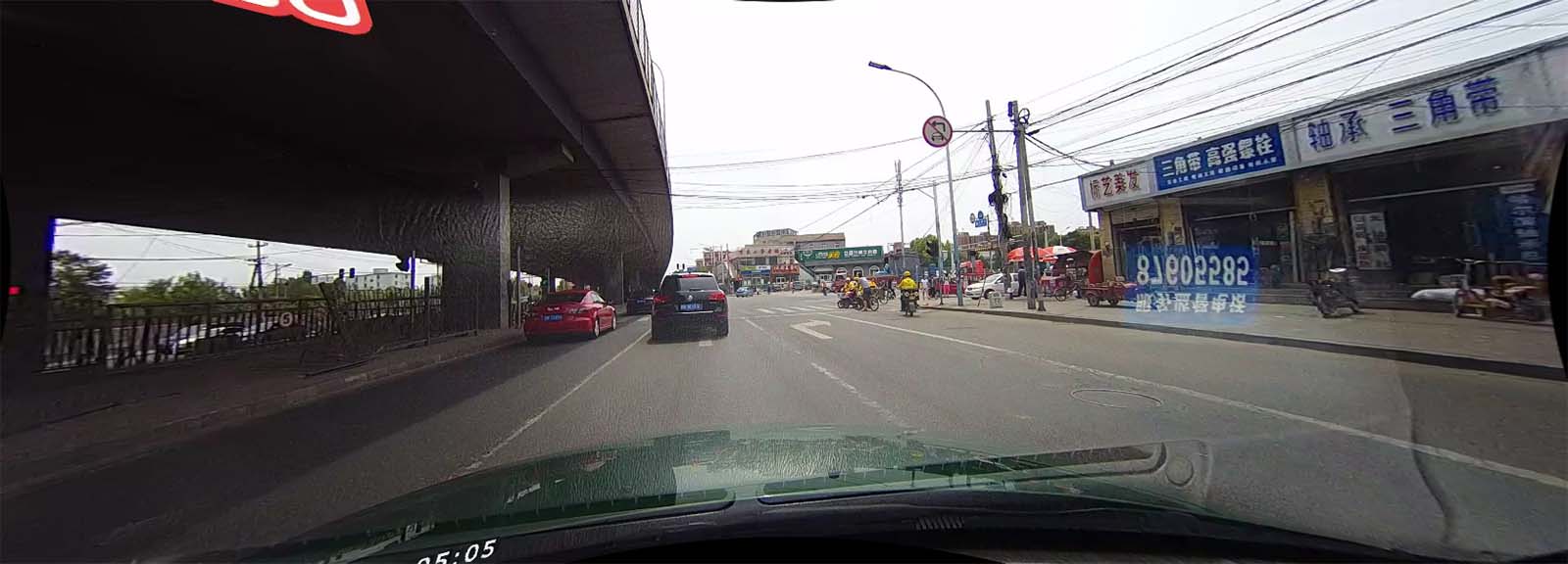} & 
\includegraphics[width=0.32\linewidth]{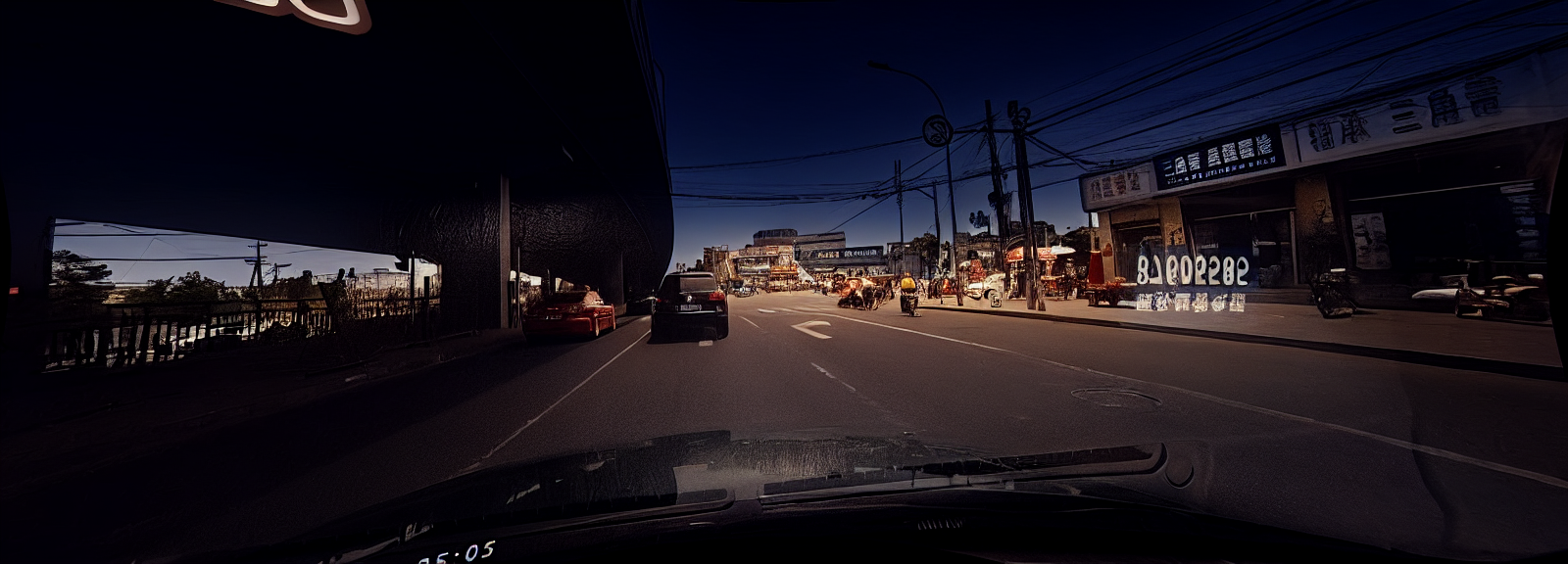} & 
\includegraphics[width=0.32\linewidth]{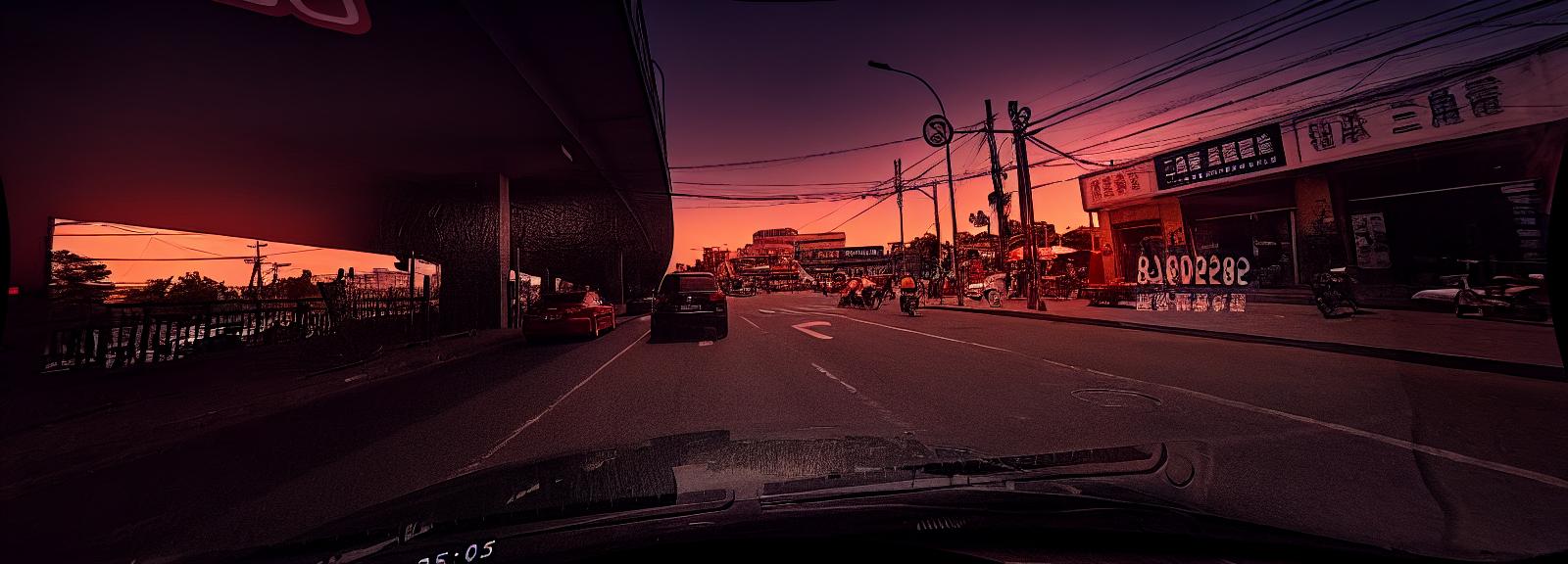} \\[-2pt]
\footnotesize Normal & \footnotesize Night & \footnotesize Dusk \\[1pt]
\includegraphics[width=0.32\linewidth]{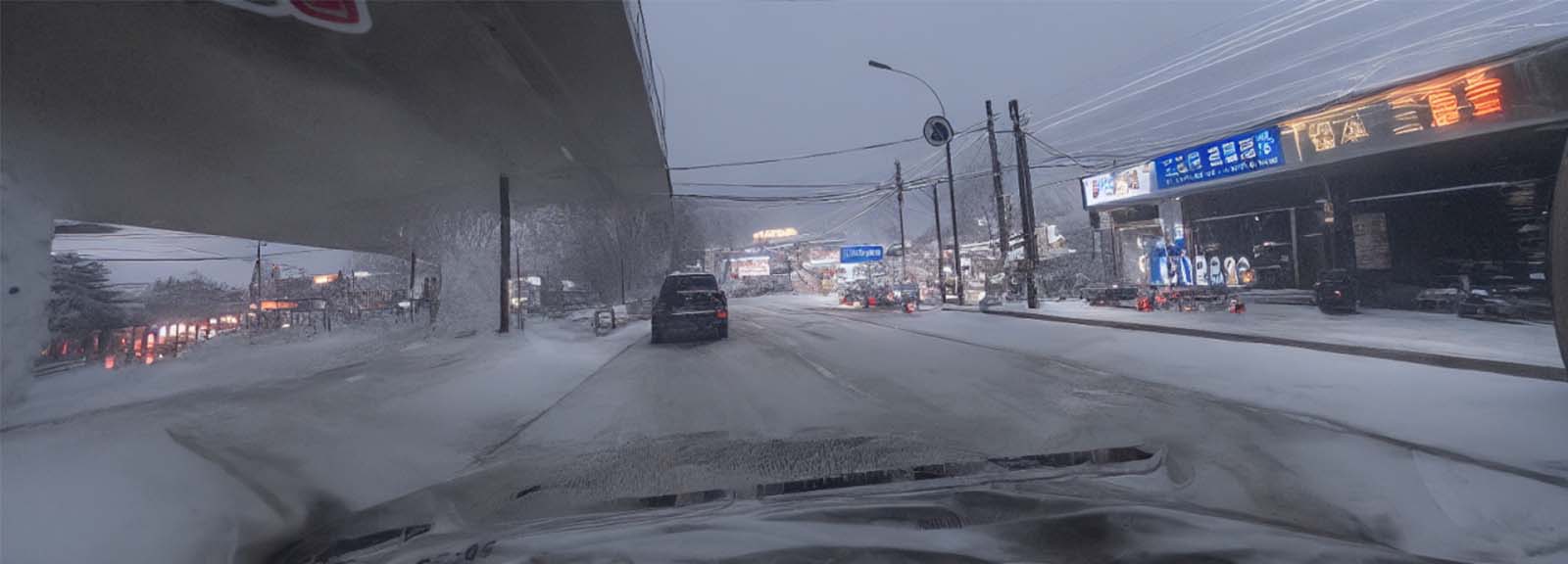} & 
\includegraphics[width=0.32\linewidth]{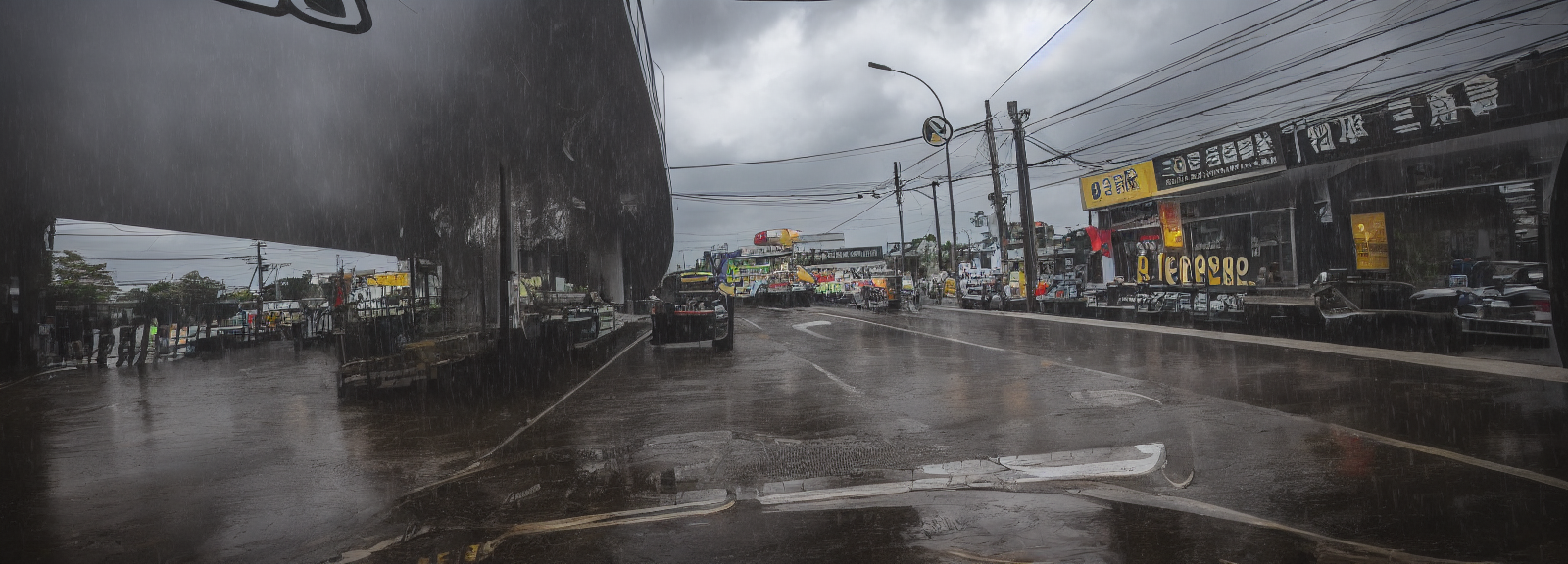} & 
\includegraphics[width=0.32\linewidth]{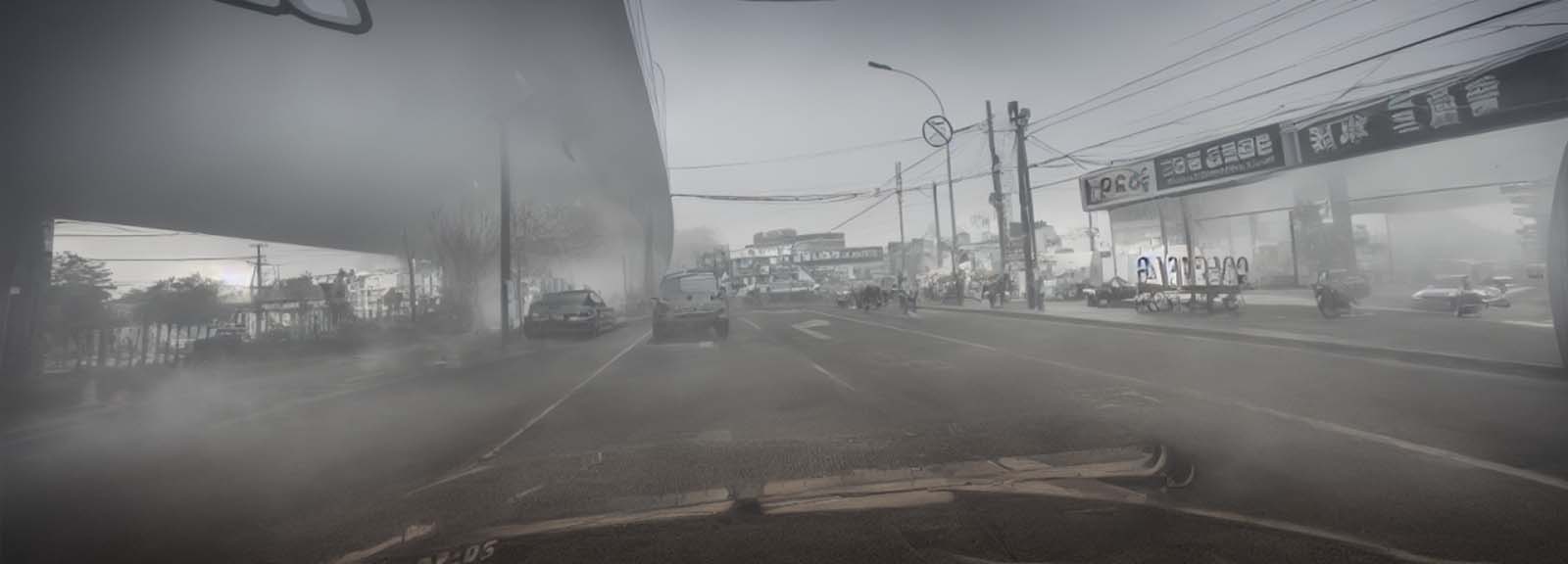} \\[-2pt]
\footnotesize Snow & \footnotesize Rain & \footnotesize Fog \\
\end{tabular}
\vspace{-1em}
\captionof{figure}{\textbf{High-fidelity weather and lighting transformations generated by our HG-Lane framework.}
Lane labels are preserved exactly, while the remaining scene semantics are kept consistent across Normal, Night, Dusk, Snow, Rain, and Fog conditions.}
\label{fig:teaser}
\end{center}
}]
\renewcommand{\thefootnote}{\fnsymbol{footnote}}
\setlength{\skip\footins}{2pt plus 1pt minus 1pt}
\footnotetext[1]{Equal contribution.} \footnotetext[2]{Corresponding author (\textit{qklee.lz@gmail.com}).}
\begin{abstract}
Lane detection is a crucial task in autonomous driving, helping to ensure the safe operation of vehicles. However, current datasets like CULane and TuSimple have relatively limited data under extreme weather conditions, such as rain, snow, and fog, which makes detection models unreliable in extreme conditions, potentially leading to serious safety-critical failures on the road. To address this, we propose \textbf{\textit{HG-Lane}}, a \textbf{H}igh-fidelity \textbf{G}eneration framework for \textbf{Lane} Scenes under adverse weather and lighting conditions, without the need for re-annotation. Based on our framework, we further propose a benchmark that includes adverse weather and lighting conditions, with 30,000 images. Experimental results demonstrate that our method consistently and significantly improves the performance of all the related lane detection networks. Taking the state-of-the-art CLRNet as an example, the overall mF1 on our benchmark increases by 20.87\%. The F1@50 for the overall, normal, snow, rain, fog, night, and dusk categories increases by 19.75\%, 8.63\%, 38.8\%, 14.96\%, 26.84\%, 21.5\%, and 12.04\%, respectively. The Code and dataset are available at \url{https://github.com/zdc233/HG-Lane}.
\end{abstract}  
\section{Introduction}
\label{sec:intro}

Lane detection is a crucial task in autonomous driving and Advanced Driver Assistance System (ADAS), which helps vehicles localize themselves better and drive more safely.

In recent years, more and more works have demonstrated their excellent performance in the lane detection task, such as UFLD \cite{pizzati2019lane}, RESA \cite{zheng2021resa}, CLRNet \cite{hu2022clrnet}, etc. However, most of the current state-of-the-art (SOTA) works are based on the CULane \cite{pan2018SCNN}, TuSimple \cite{pizzati2019lane} and LLAMAS \cite{llamas2019} datasets, the vast majority of whose samples are sunny and daytime scenes, lacking samples of adverse weather conditions. This deficiency causes a significant drop in the performance of detection models under adverse weather conditions, which affects the safety of autonomous driving.

Although it is possible to collect these data manually, the low frequency of adverse weather occurrences and the high cost of re-annotation make it inefficient and difficult to build in a short period of time. Besides, it should be noted that some existing works can achieve image restoration, such as WGWS-Net \cite{Zhu_2023_CVPR}, Domain Translation Multi-weather Restoration \cite{Patil_2023_ICCV}, and CycleGAN \cite{shao2021cyclegan}, but they require retraining for each type of weather condition and the process of image restoration adds extra overhead, making it difficult to meet the requirements for real-time performance. In addition to these methods, there are also works based on Diffusion models that achieve realistic data generation, such as WeatherDG \cite{qian2025weatherdg} and DA-Fusion \cite{tra2024dafusion}.
However, they still require re-annotation of data and fine-tuning of the diffusion model, which incurs additional costs.

In this paper, we present \textbf{\textit{HG-Lane}}, a \textbf{H}igh-fidelity \textbf{G}enerative framework that generates realistic \textbf{Lane} scenes under adverse weather and lighting conditions—such as snow, rain, fog, night, and dusk—\textit{without requiring any re-annotation}. 
Specifically, we extract the normal-condition images from the CULane dataset and preprocess them with a semantic-preserving strategy that fuses Canny edges and lane annotations. These are used to guide the generative model via a dual-stage pipeline: ControlNet (Canny) handles structure preservation, while ControlNet (InstructPix2Pix) refines lighting and style. By applying category-specific prompts, we successfully generate realistic lane scenes across five new adverse-condition categories.

Building upon HG-Lane, we further introduce a new benchmark consisting of 30,000 generated images across six conditions, significantly enriching the diversity of lane detection scenarios. We evaluate multiple state-of-the-art lane detection models on this benchmark and conduct comprehensive ablation studies to validate the effectiveness of each component in our framework. We also compare multiple generative models and suppression modules, and carry out tests on real-world data.
Experimental results demonstrate that the proposed method stably and significantly improves the detection accuracy of all relevant networks, exhibiting strong generalization capability and structural soundness across various settings.

The main contributions can be summarized as follows:

\begin{itemize}
    \item We propose HG-Lane, a novel framework that generates high-fidelity lane scenes under diverse weather and lighting conditions, while \textbf{preserving lane semantics and requiring no re-annotation in a consistent manner}.
    \vspace{0.5em}
    \item We design a \textbf{dual-stage} generation strategy using ControlNet with Canny and InstructPix2Pix guidance, enhanced by a semantic-aware preprocessing pipeline to retain lane structure and scene consistency.
    \vspace{0.5em}
    \item We construct a new benchmark containing \textbf{30,000} lane images across six categories \emph{(normal, snow, rain, fog, night, dusk)}, enabling systematic evaluation of model robustness under challenging conditions.
    \vspace{0.5em}
    \item Extensive experiments show that our method improves lane detection accuracy across multiple models. Taking SOTA CLRNet as an example, it increases overall mF1 by 20.87\%, with significant gains in all subcategories: snow \textbf{\emph{(+38.8\%)}}, night \textbf{\emph{(+21.5\%)}}, fog \textbf{\emph{(+26.84\%)}}, etc.
\end{itemize}
\section{Related Work}
\subsection{Generative Models}
Generative models have always attracted significant attention and have seen rapid development in recent years. Early methods, such as \ding{68} \textbf{Variational Autoencoders (VAEs)}\cite{die2014vae}, \ding{68} \textbf{Generative Adversarial Networks (GANs)}\cite{goodfellow2014gan}, and \ding{68}\textbf{ Flow-Based models} \cite{rezende2015flow}, have shown potential in generating images but also have certain limitations. For instance, GANs are not good at generating samples they have not observed, while VAEs and Flow-Based models are not proficient at generating high-fidelity samples. In recent years, many successes in photorealistic image generation have been achieved by Diffusion models \cite{ho2020ddpm} \cite{alex2021improveddpm} \cite{saharia2022text2img} \cite{alex2022glide} \cite{ramesh2022text-condi}. Compared to GANs, \ding{68} \textbf{Diffusion models} have been proven to generate higher-fidelity samples \cite{dhariwal2021difvsgan}, and developments such as classifier-free guidance \cite{ho2022classifier-free} have made text-to-image generation possible.

\subsection{Condition Control}
Condition control in diffusion models has always been a research hotspot. Stable Diffusion \cite{ro2022sd}, trained on the LAION-5B \cite{sc2022laion5b} dataset and based on CLIP \cite{ra2021clip}, can achieve text-guided image generation. \ding{68} \textbf{ControlNet} \cite{he2023controlnet} fine-tunes Stable Diffusion and uses multiple conditions to guide image generation, including text prompts and edge maps, which can better preserve the overall semantics of the image. \ding{68} \textbf{InstructPix2Pix} \cite{dong2023instructp2p}, on the other hand, can directly edit the original image and retain more details. Currently, combining multiple ControlNets together has shown powerful effects. For example, ComfyUI\cite{comfyui} provides a convenient workflow to achieve this in a highly flexible manner.

\subsection{Lane Detection}
Lane detection methods can be categorized into several approaches. Segmentation-based approaches like SCNN \cite{pan2018SCNN} and RESA \cite{zheng2021resa}, utilize pixel classification and post-clustering strategies. Row-wise classification methods such as CondLaneNet \cite{liu2021condlanenet} enhance computational efficiency by dividing the input image into grids and predicting lanes row by row. Anchor-based methods, including UFLD \cite{qin2020ufld} and \ding{68} \textbf{CLRNet} \cite{hu2022clrnet}, employ various anchor strategies to improve detection of lanes. Parametric prediction methods, represented by PolyLaneNet \cite{tabelini2021polylanenet} and LSTR \cite{liu2021lstr}, achieve end-to-end detection by modeling lane lines as curve equations. Each approach contributes uniquely to advancing lane detection capabilities within modern autonomous driving.

\begin{figure*}[t]
    \centering
    \includegraphics[width=0.95\textwidth]{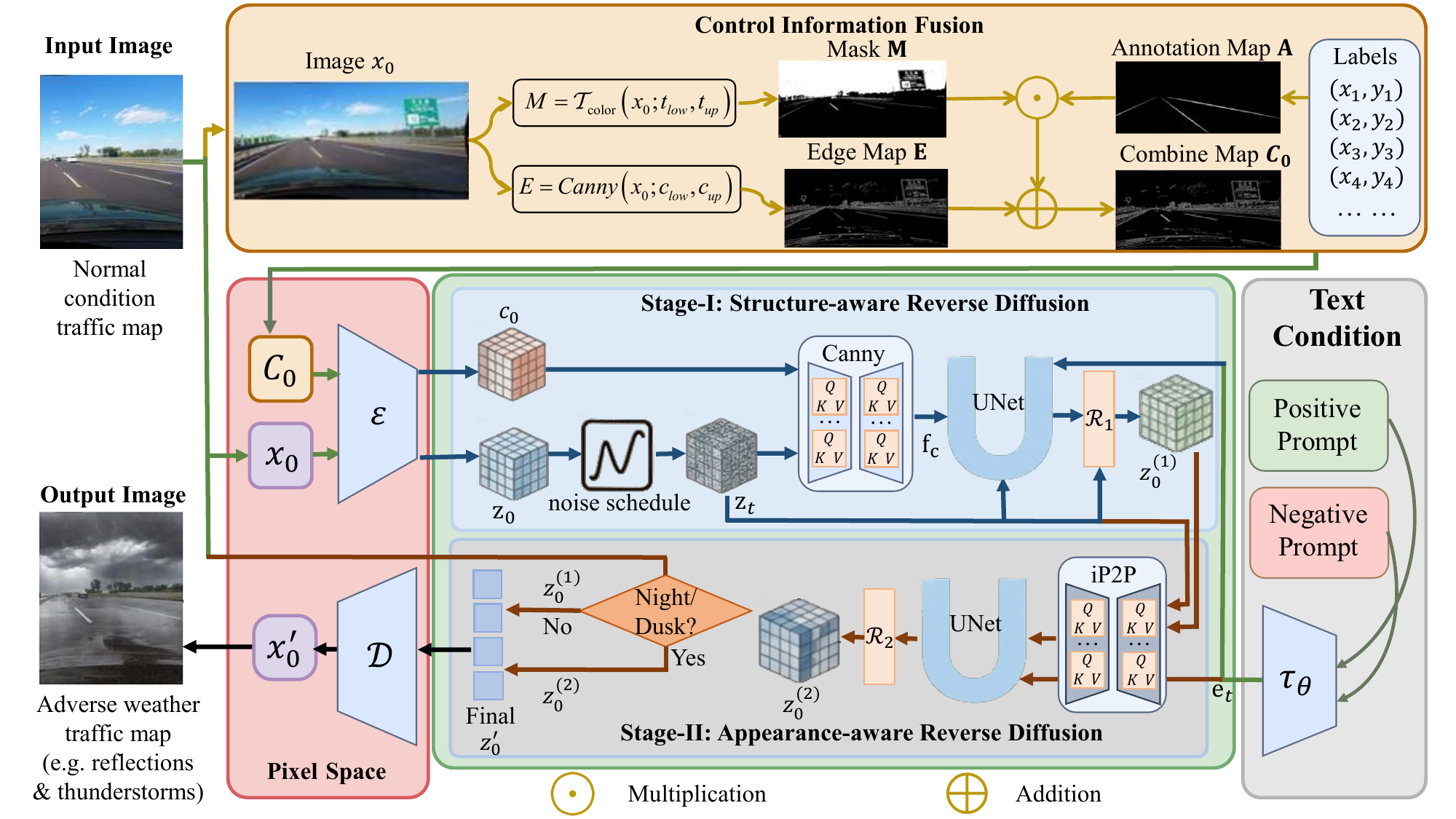}
    \caption{
        \textbf{Overview of the proposed HG-Lane.}
        The input image is first processed into a fused control map combining color-based masks, Canny edges and lane annotations.
        In Stage-I, a Canny-ControlNet enforces lane geometry during reverse diffusion in latent space.
        In Stage-II, an InstructPix2Pix-ControlNet optionally refines appearance for ``night'' and ``dusk''.
        Finally, the latent is decoded back to pixel space. All components are pretrained; no fine-tuning is required.
    }
    \label{fig:framework}
\end{figure*}
\section{Method}
HG-Lane is a dual-stage, control-guided diffusion framework that generates lane images with diverse weather conditions and illumination conditions while preserving lane geometry. 
Instead of fine-tuning diffusion models, HG-Lane reuses pretrained components and injects task-specific structure via control signals. 
As shown in Figure ~\ref{fig:framework}, the pipeline consists of four components:
(1) \textbf{Control Information Fusion} that fuses lane annotations, color masks, and Canny edges into a unified control map;
(2) \textbf{Stage-I Structure-aware Reverse Diffusion} driven by a Canny-based ControlNet to enforce lane geometry;
(3) \textbf{Stage-II Appearance-aware Reverse Diffusion} using an InstructPix2Pix ControlNet to refine style for selected categories (night/dusk);
and (4) \textbf{Pixel-space Conversion} that decodes the latent back to the image domain.

We denote the pretrained VAE encoder and decoder as
$\mathcal{E}(\cdot)$ and $\mathcal{D}(\cdot)$,
and the pretrained latent diffusion model (UNet backbone plus ControlNet) as
$\mathcal{U}_{\phi}$, with all parameters kept frozen during our pipeline.

\vspace{-0.2em}
\subsection{Control Information Fusion}
\vspace{-0.2em}
Let the input lane image be $\mathbf{x}_0 \in \mathbb{R}^{H\times W\times 3}$ and the corresponding lane annotation map (extracted from the original labels and filled to a certain width to ensure they cover the area where the lane lines are located) be $\mathbf{A} \in \{0,1\}^{H\times W}$.
The goal of this stage is to construct a control map $\mathbf{C}_0$ that encodes both lane geometry and local edge cues while being robust to dashed or incomplete markings.
\paragraph{Geometric priors from color and edges.}
We first apply a color-thresholding operator to extract lane-related regions:
\begin{equation}
\mathbf{M} 
= \mathcal{T}_{\text{color}}
\bigl(\mathbf{x}_0;\,\boldsymbol{t}_{\text{low}},\boldsymbol{t}_{\text{up}}\bigr),
\qquad 
\mathbf{M} \in \{0,1\}^{H\times W},
\end{equation}
where $\boldsymbol{t}_{\text{low}}$ and $\boldsymbol{t}_{\text{up}}$ are lower and upper bounds of the color threshold and $\mathbf{M}$ denotes the resulting mask map.
In parallel, we compute the edge map using the Canny operator:
\begin{equation}
\mathbf{E} 
= \operatorname{Canny}
\bigl(\mathbf{x}_0;\,\boldsymbol{c}_{\text{low}},\boldsymbol{c}_{\text{up}}\bigr),
\end{equation}
where $\boldsymbol{c}_{\text{low}}$ and $\boldsymbol{c}_{\text{up}}$ are lower and upper bounds of the canny threshold.
\vspace{-0.5em}
\paragraph{Fusion with annotations.}
To avoid over-relying on dense, continuous annotations (which may conflict with dashed lane markings), we combine them with $\mathbf{M}$ and $\mathbf{E}$ via:
\begin{equation}
\mathbf{C}_0 
= \bigl(\mathbf{A} \odot \mathbf{M}\bigr) \;\oplus\; \mathbf{E},
\end{equation}
where $\odot$ denotes element-wise multiplication and $\oplus$ denotes pixel-wise logical OR (or max).  
Here, $\mathbf{A} \odot \mathbf{M}$ suppresses annotation pixels outside the color-consistent region, while $\mathbf{E}$ injects additional edge cues where Canny succeeds even if annotations are sparse or imperfect.
Our ablation shows that using $\mathbf{E}$ alone leads to missing or misaligned lanes in complex conditions, whereas $(\mathbf{A} \odot \mathbf{M}) \oplus \mathbf{E}$ yields more reliable lane structure priors.

\subsection{Stage-I: Structure-aware Reverse Diffusion}

The first reverse diffusion stage enforces lane geometry in the latent space using the fused control map $\mathbf{C}_0$ and a Canny-based ControlNet.

\paragraph{Latent encoding and forward diffusion.}
We first encode the control map and the original image into the latent space:
\begin{equation}
\mathbf{c}_0 = \mathcal{E}(\mathbf{C}_0), 
\qquad
\mathbf{z}_0 = \mathcal{E}(\mathbf{x}_0),
\end{equation}
where $\mathbf{c}_0, \mathbf{z}_0 \in \mathbb{R}^{H'\times W'\times d}$.
A standard forward diffusion process gradually corrupts $\mathbf{z}_0$ into $\mathbf{z}_t$:
\begin{equation}
q(\mathbf{z}_t \mid \mathbf{z}_0)
= \mathcal{N}\!\left(\mathbf{z}_t;
    \sqrt{\bar{\alpha}_t}\,\mathbf{z}_0,\,
 1-\bar{\alpha}_t
\right),
\  t=1,\dots,T,
\end{equation}
where $\{\bar{\alpha}_t\}$ is the cumulative noise schedule and $\mathbf{z}_T$ is sampled at the end of the forward process.

\paragraph{Canny-ControlNet guided denoising.}
Conditioned on $\mathbf{c}_0$ and text prompts, Stage-I reverse diffusion reconstructs a geometry-consistent latent $\mathbf{z}_0^{(1)}$.
We use a Canny-trained ControlNet, denoted as $\mathcal{C}_{\text{canny}}$, together with a text encoder $\tau_{\theta}$ that maps positive/negative prompts to conditional embeddings $\mathbf{e}_t$:
\begin{equation}
\mathbf{e}_t = \tau_{\theta}(\text{prompt}), 
\qquad
\mathbf{f}_c = \mathcal{C}_{\text{canny}}(\mathbf{c}_0, \mathbf{z}_t).
\end{equation}
At each timestep $t$, the UNet backbone predicts the noise using both control features $\mathbf{f}_c$ and prompt embedding $\mathbf{e}_t$:
\begin{equation}
\hat{\boldsymbol{\epsilon}}_t^{(1)} 
= \mathcal{U}_{\phi}\bigl(\mathbf{z}_t, t, \mathbf{e}_t, \mathbf{f}_c\bigr),
\end{equation}
and the reverse update follows the standard DDPM/DDIM rule
\begin{equation}
\mathbf{z}_{t-1} = \mathcal{R}_1\bigl(\mathbf{z}_t, \hat{\boldsymbol{\epsilon}}_t^{(1)}\bigr),
\end{equation}
where $\mathcal{R}_1$ denotes one denoising step in Stage-I.
After $T$ steps, we obtain the Stage-I latent:
\begin{equation}
\mathbf{z}_0^{(1)} 
= \mathcal{R}_1^{(T)}\bigl(\mathbf{z}_T,\hat{\boldsymbol{\epsilon}}_t^{(1)}\bigr).
\end{equation}
This stage focuses on enforcing lane geometry and structural consistency under the Canny control.

\paragraph{Cross-attention view.}
Inside the Attention-UNet of $\mathcal{U}_{\phi}$, the prompt embedding 
$\mathbf{e}_t \in \mathbb{R}^{n \times c}$ 
and latent feature 
$\mathbf{z}_t \in \mathbb{R}^{(h \times w) \times c}$ 
interact through cross-attention:
\begin{equation}
\mathbf{Q} = \mathbf{z}_t \mathbf{W}^{Q}, \qquad
\mathbf{K} = \mathbf{e}_t \mathbf{W}^{K}, \qquad
\mathbf{V} = \mathbf{e}_t \mathbf{W}^{V},
\end{equation}
\begin{equation}
\operatorname{Attn}(\mathbf{z}_t,\mathbf{e}_t) 
= \operatorname{Softmax}\!\left(
    \frac{\mathbf{Q}\mathbf{K}^\top}{\sqrt{d}}
\right)\mathbf{V},
\end{equation}
where $\mathbf{W}^{Q},\mathbf{W}^{K},\mathbf{W}^{V}\in\mathbb{R}^{c\times d}$ are learned projections.
This mechanism allows lane-specific prompts to modulate lane structures generated under Canny control condition.

\subsection{Stage-II: Appearance-aware Reverse Diffusion}

While Stage-I constrains lane structure, different weather or illumination categories require distinct appearance styles (e.g., night, dusk, snow, rain, fog).  
Stage-II introduces an InstructPix2Pix (iP2P) ControlNet to refine appearance in a targeted manner.

\paragraph{Category-dependent refinement.}
For categories such as ``night'' and ``dusk'', solely using Canny control tends to preserve structure but fails to fully capture desired lighting and global tone.
We therefore introduce a second reverse diffusion stage with an iP2P ControlNet, denoted $\mathcal{C}_{\text{iP2P}}$, which refines from $\mathbf{z}_0^{(1)}$ toward an appearance-enhanced latent $\mathbf{z}_0^{(2)}$:
\begin{equation}
\hat{\boldsymbol{\epsilon}}_t^{(2)} 
= \mathcal{U}_{\phi}\bigl(\mathbf{z}_t, t, \mathbf{e}_t, \mathcal{C}_{\text{iP2P}}\bigl(\mathbf{z}_0^{(1)}, \mathbf{z}_t\bigr)\bigr),
\end{equation}
\begin{equation}
\mathbf{z}_0^{(2)} 
= \mathcal{R}_2^{(T)}\bigl(\mathbf{z}_T,\hat{\boldsymbol{\epsilon}}_t^{(2)}\bigr),
\end{equation}
where $\mathcal{R}_2$ denotes the second-stage denoising process and $\mathbf{e}_t$ describes the desired style (e.g., night scene with illuminated lanes).

For categories such as ``snow'', ``rain'' and ``fog'', we empirically observe that a single Canny ControlNet already provides satisfactory visual realism. Therefore, for these categories, we skip Stage-II and simply set
\begin{equation}
\mathbf{z}_0' = \mathbf{z}_0^{(1)}.
\end{equation}
For categories ``night'' and ``dusk'', we set
\begin{equation}
\mathbf{z}_0' = \mathbf{z}_0^{(2)}.
\end{equation}

\subsection{Pixel-space Conversion and Sampling Strategy}

In the final step, the refined latent $\mathbf{z}_0'$ is mapped back to pixel space using the pretrained VAE decoder:
\begin{equation}
\mathbf{x}_0' = \mathcal{D}(\mathbf{z}_0'),
\end{equation}
yielding the generated lane image under the target weather/illumination condition.

It is worth emphasizing that the entire HG-Lane framework is \emph{training-free}: all VAEs, UNet backbones, and ControlNets are pretrained and kept frozen.  
In practice, we can sample multiple outputs using different random seeds:
\begin{equation}
\mathbf{x}_0'^{(k)} = \mathcal{D}\bigl(\mathbf{z}_0'^{(k)}\bigr),
\qquad k = 1,\dots,K,
\end{equation}
evaluate them with metrics such as F1@50 and FID, and select the seed that provides the best trade-off between geometric accuracy and visual realism for deployment.
This makes HG-Lane readily integrable into existing lane detection pipelines while providing controllable, high-fidelity synthetic data across a wide range of conditions.

\begin{table*}[t]
\centering
\caption{\textbf{Performance on CLRNet (ResNet-18).} For Normal, Snow, Rain, Fog, Night, and Dusk, the metric used is F1@50.}
\small
\setlength{\belowrulesep}{0pt}
\setlength{\tabcolsep}{2.3mm}
\begin{tabular}{l|cccccc|ccc}
    \Xhline{1.2pt}
    \rowcolor{violet!15} Method &Normal&Snow&Rain&Fog&Night&Dusk&F1@50&F1@75&mF1\\
    \Xhline{1.2pt}
    w/o aug (4k)      &83.01&46.08&70.34&58.54&70.09&79.95&68.74&46.39&42.16\\
    w/o aug (4k*6)    &87.28&51.79&71.48&58.31&62.67&75.62&68.89&50.87&45.42\\
   \rowcolor{green!10} \textbf{ w/ aug (4k*6)}     
                        &\textbf{91.64}&\textbf{84.88}&\textbf{85.30}&
                        \textbf{85.38}&\textbf{91.59}&\textbf{91.99}&
                        \textbf{88.49}&\textbf{72.79}&\textbf{63.03}\\
    \Xhline{1.2pt}
\end{tabular}
\label{table:Comparison}
\end{table*}

\section{Experiment}
\subsection{Benchmark}
The normal category in our dataset was derived from the normal category in the CULane dataset, and we obtained 5,000 images via systematic sampling. The other five categories (snow, rain, fog, night, dusk) were generated by our framework, each containing 5,000 images. In total, the dataset consists of 30,000 images. The training set, validation set, and test set are divided in a 7:1:2 ratio. 



\subsection{Metrics}
We follow the evaluation protocol of the CULane dataset.
A predicted lane point is matched to the ground truth if the Intersection over Union (IoU), computed by dilating both by 30 pixels, exceeds a threshold~$\alpha$.
True positives, false positives, and false negatives are counted to compute F1.
In practice, $\alpha$ is set to 0.5 and 0.75, reported as F1@50 and F1@75, respectively. 
We also compute the mean F1 (mF1) by averaging F1 scores over $\alpha \in [0.5, 0.95]$ with a step of 0.05. 
For the Normal, Snow, Rain, Fog, Night, and Dusk scenarios, the primary metric is F1@50 in the reported results.

\subsection{Experimental Details}
\textbf{Generate Images by Our Framework.} Our framework employs ComfyUI for workflow design. The experiment is conducted on an NVIDIA GeForce RTX 3090 GPU with 24 GB of memory, running the Ubuntu 24.04 operating system. The programming language used is Python 3.8.0, and the PyTorch 1.13.1 framework is employed.  Reverse Diffusion 1 and 2 used the Euler sampler and the Karras scheduler, with a sampling step size of 30 and cfg = 6.0. 

\noindent\textbf{Lane Detection Experiment.} Using our benchmark, we tested SOTA and highly cited lane detection models from recent years. We employed the environment required by the official code of each model. Given the wide range of hardware requirements, we conducted the tests on multiple servers with different specifications and used the best model to test.

\subsection{Comparison Experiment}
We utilized CLRNet (ResNet-18) to test both the pre-augmented and post-augmented datasets. The test sets consist of 6 categories (normal, snow, rain, fog, night, dusk), each with 1,000 images. To control for the effect of dataset size on the results, before augmentation, we used 3,500 training + 500 validation normal images as one dataset, and another dataset with 6 × (3,500 training + 500 validation) normal images. After augmentation, we used datasets with 6 categories, each containing 3,500 training + 500 validation images. This ensures a consistent number of normal images used for training and validation, as well as a consistent total number of images.

The experimental results are shown in Table \ref{table:Comparison}. 
We observe that F1@50 for the normal, snow, rain, fog, night, and dusk categories increased by 8.63\%, 38.8\%, 14.96\%, 26.84\%, 21.5\%, and 12.04\%, with an overall improvement of 19.75\%. 
This validates that the data we generated addresses the robustness issue of the model under adverse weather and lighting conditions.

\subsection{Performance on various baselines}

\begin{figure*}[h]
\centering
\includegraphics[width=1\linewidth]{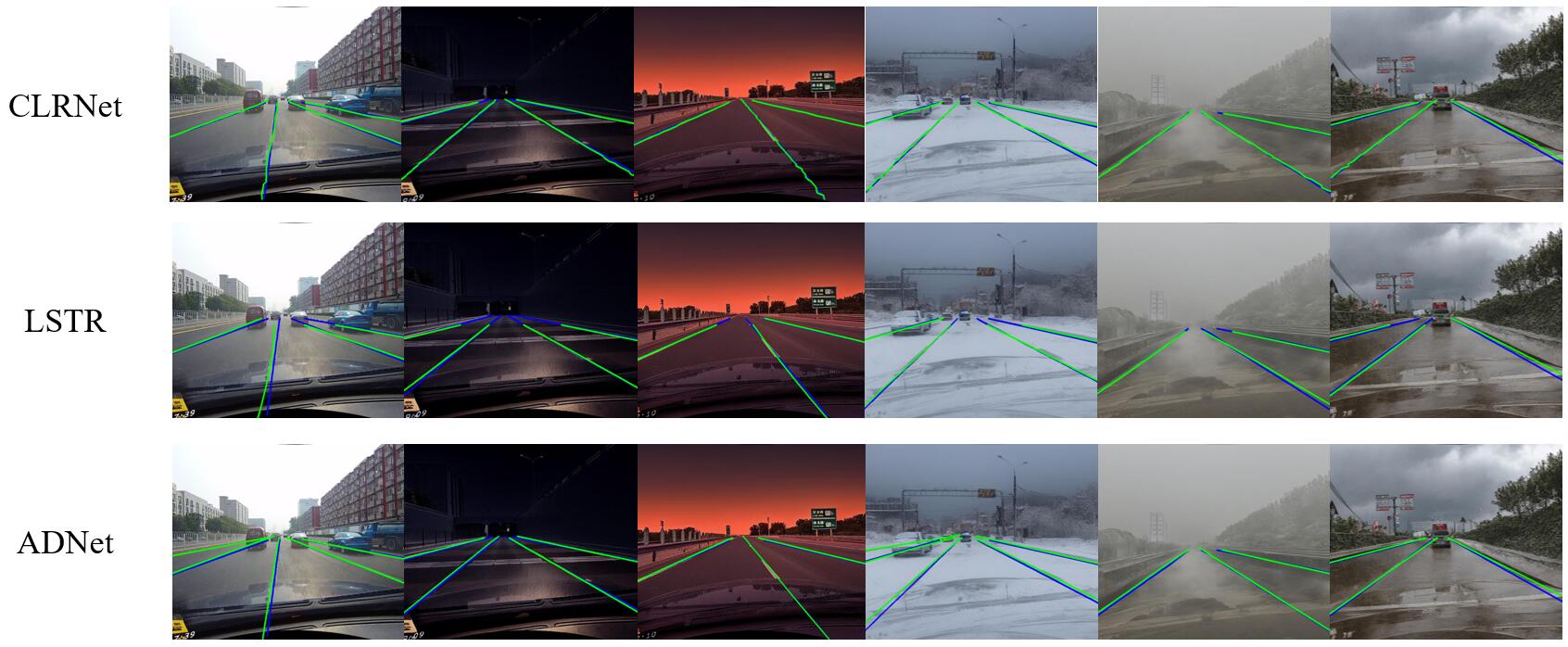}
\caption{\textbf{Results of some baselines.} The green lines in the figure represent the predicted values, while blue lines represent ground truth.}
\label{fig:results}
\end{figure*}

As shown in Table \ref{table:baseline}, we conducted tests on the state-of-the-art (SOTA) and highly cited lane detection baselines in recent years. All our training configurations are based on the open-source code of the respective baselines, with only the same epoch set to 12 and without using any pre-trained weights. The results demonstrate that our dataset exhibits good compatibility across all the baselines under diverse evaluation settings. As shown in Figure \ref{fig:results}, we present partial results of the baselines on six categories. The green lines represent the predicted values, while the blue lines represent the ground truth. It can be observed that the two lines fit well with each other. We specifically list ADNet, and it can be found that it also predicts the leftmost lane line, even though this lane line is not labeled in the CULane dataset.


\subsection{Ablation Study}

\begin{table}[h]
    \centering
    \caption{\textbf{Ablation study.} The symbol \checkmark indicates whether it is used, while "before" and "after" denote the sequence of usage.}
    \setlength{\belowrulesep}{0pt}
    \setlength{\tabcolsep}{1.5mm}
    \begin{tabular}{ccccccc}
        \Xhline{1.2pt}
        \rowcolor{violet!15} \#&Mask&Canny&iP2P&F1@50&F1@75&mF1\\
        \Xhline{1.2pt}
        \rowcolor{green!10} 1&\checkmark&\checkmark{before}&\checkmark{after}&\textbf{88.49}&\textbf{72.79}&\textbf{63.03}\\        
        2&\checkmark&\checkmark{after}&\checkmark{before}&62.56&41.01&39.44\\
        \rowcolor{gray!10}
        3&\checkmark&&\checkmark&59.75&40.33&39.33\\
        4&\checkmark&\checkmark&&64.01&43.78&40.12\\
        \rowcolor{gray!10}
        5&&\checkmark{before}&\checkmark{after}&86.81&70.12&61.35\\
        \Xhline{1.2pt}
    \end{tabular}
    \label{table:ablation}
\end{table}

\begin{table*}[t]
\centering
\small
\setlength{\belowrulesep}{1pt}
\setlength{\tabcolsep}{2.1mm}

\caption{
\textbf{Performance comparison on CULane benchmarks.}
For Normal, Snow, Rain, Fog, Night, and Dusk, the metric is F1@50 to provide a clearer comparison across different environmental conditions. Best results are in \textbf{bold}, second best are \underline{underlined}.}

\begin{tabular}{l|cccccc|ccc}
    \Xhline{1.2pt}
    \rowcolor{violet!12}
    Method & Normal & Snow & Rain & Fog & Night & Dusk & F1@50 & F1@75 & mF1 \\
    \Xhline{1.2pt}

    \rowcolor{gray!10}
    \multicolumn{10}{c}{\textbf{Classical / Light Backbones}} \\
    \Xhline{1.2pt}
    SCNN (ResNet-34)\cite{pan2018SCNN} & 75.91 & 68.96 & 67.56 & 62.54 & 70.51 & 70.46 & 72.97 & 31.34 & 20.08\\
    SAD (ENet)\cite{hou2019sad}        & 76.75 & 64.71 & 68.85 & 61.73 & 73.39 & 76.68 & 70.02 & 30.91 & 19.36 \\
    LSTR (ResNet-34)\cite{liu2021lstr} & 84.06 & 74.67 & 76.61 & 74.40 & 83.71 & 84.38 & 65.07 & 59.75 & 16.48\\

    \Xhline{1.2pt}
    \rowcolor{gray!10}
    \multicolumn{10}{c}{\textbf{UFLD / RESA Family}} \\
    \Xhline{1.2pt}
    UFLD (ResNet-18)\cite{qin2020ufld} & 69.59 & 59.29 & 60.77 & 56.52 & 59.69 & 63.97 & 61.65 & 17.56 & 13.42\\
    UFLD (ResNet-34)\cite{qin2020ufld} & 74.42 & 62.52 & 66.28 & 64.72 & 72.77 & 74.01 & 69.14 & 18.99 & 14.61\\
    RESA (ResNet-18)\cite{zheng2021resa} & 78.44 & 69.28 & 70.17 & 69.41 & 77.34 & 78.47 & 73.86 & 28.29 & 19.85\\
    RESA (ResNet-34)\cite{zheng2021resa} & 77.31 & 68.90 & 69.20 & 69.12 & 77.42 & 77.71 & 73.38 & 27.58 & 19.36\\

    \Xhline{1.2pt}
    \rowcolor{gray!10}
    \multicolumn{10}{c}{\textbf{LaneATT / LaneAF / CondLaneNet Family (Modern CNN Baselines)}} \\
    \Xhline{1.2pt}
    LaneATT (ResNet-18)\cite{lu2021laneatt} & 81.78 & 72.55 & 73.17 & 71.48 & 80.53 & 80.75 & 76.75 & 44.71 & 34.02\\
    LaneATT (ResNet-34)\cite{lu2021laneatt} & 81.92 & 72.77 & 73.34 & 71.49 & 80.63 & 80.88 & 76.99 & 45.01 & 34.18\\
    LaneAF (ERFNet)\cite{hala2021LaneAF} & 89.63 & 84.98 & 84.56 & 84.78 & 89.12 & 89.63 & 86.66 & 67.01 & 58.78\\
    LaneAF (DLA-34)\cite{hala2021LaneAF} & 90.63 & 85.15 & 84.94 & 85.19 & 90.02 & 90.21 & 87.70 & 67.72 & 59.98\\
    CondLaneNet (ResNet-18)\cite{liu2021condlanenet} & 90.58 & 83.52 & 84.15 & 84.59 & 90.04 & 89.89 & 88.44 & 68.76 & 59.05\\
    CondLaneNet (ResNet-34)\cite{liu2021condlanenet} & 91.05 & 84.35 & 85.31 & 84.98 & 90.56 & 90.14 & 88.53 & 69.56 & 59.87\\

    \Xhline{1.2pt}
    \rowcolor{gray!10}
    \multicolumn{10}{c}{\textbf{CLR / CLRKD / GANet Family (State-of-the-art CNN Lanes)}} \\
    \Xhline{1.2pt}
    GANet (ResNet-18)\cite{wang2022GANet} & 93.22 & 88.01 & 88.55 & 88.34 & 92.59 & \textbf{94.36} & 90.88 & 75.31 & 66.18\\
    GANet (ResNet-34)\cite{wang2022GANet} & \underline{93.96} & \underline{88.95} & \underline{88.89} & \underline{88.78} & \underline{93.30} & \underline{93.54} & \underline{91.25} & 75.55 & \underline{66.89}\\
    CLRNet (ResNet-18)\cite{hu2022clrnet} & 91.64 & 84.88 & 85.30 & 85.38 & 91.59 & 91.99 & 88.49 & 72.79 & 63.03\\
    CLRNet (ResNet-34)\cite{hu2022clrnet} & 91.25 & 80.96 & 85.24 & 83.15 & 91.15 & 91.84 & 87.38 & 70.76 & 60.89\\
    ADNet (ResNet-18)\cite{Xiao2023adnet} & 81.03 & 70.76 & 74.29 & 72.11 & 81.12 & 82.09 & 76.98 & 51.59 & 42.33\\
    ADNet (ResNet-34)\cite{Xiao2023adnet} & 81.89 & 71.20 & 75.89 & 73.33 & 81.89 & 82.65 & 77.37 & 52.02 & 42.84\\
    CLRerNet\cite{hirote2024CLRerNet} & 92.34 & 82.54 & 85.47 & 85.33 & 91.54 & 91.88 & 90.68 & \underline{76.27} & 65.31\\
    CLRKDNet (ResNet-18)\cite{qi2024CLRKDNet} & 92.15 & 85.33 & 85.46 & 85.52 & 92.02 & 92.69 & 88.53 & 73.11 & 64.01 \\
    CLRKDNet (DLA34)\cite{qi2024CLRKDNet} & 92.52 & 85.62 & 85.74 & 85.33 & 91.99 & 93.21 & 88.92 & 73.49 & 64.75\\

    \textbf{FENet}\cite{wang2024FENet} 
      & \textbf{95.78} & \textbf{89.47} & \textbf{90.28} & \textbf{95.05} 
      & \textbf{95.68} & 90.43 
      & \textbf{92.76} & \textbf{82.17} & \textbf{69.66} \\

    \Xhline{1.2pt}
\end{tabular}
\label{table:baseline}
\end{table*}

\begin{figure*}[t]
\centering
\includegraphics[width=1\textwidth]{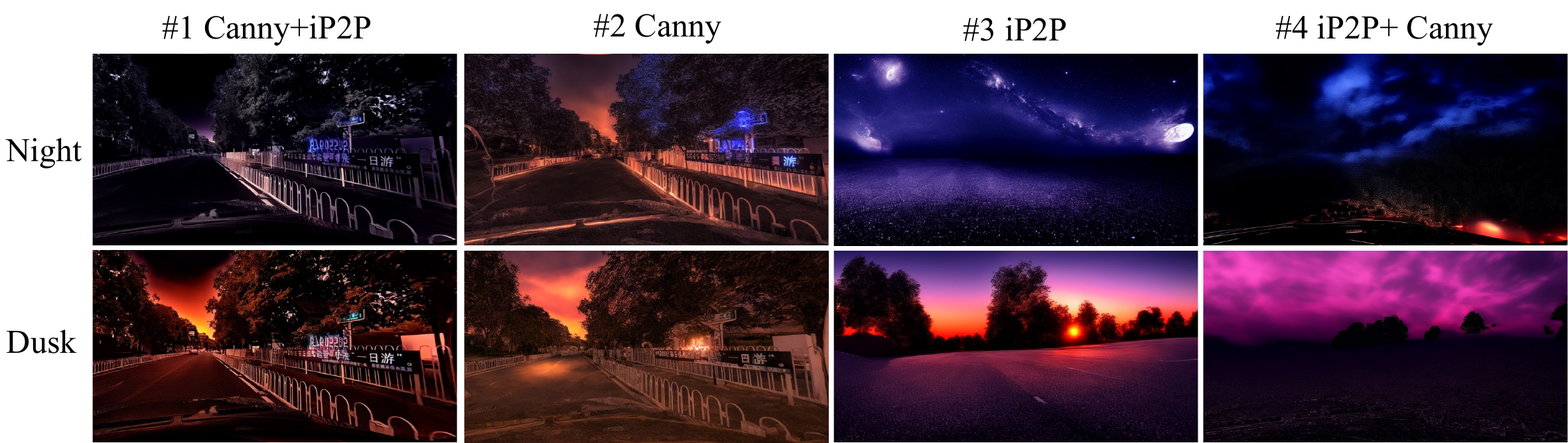}
\caption{\textbf{Ablation Study.} Experiment \#1, \#2, \#3, and \#4 demonstrate the generation results of using Canny or InstructPix2Pix individually, as well as the effects of their combined order.}
\label{fig:ablation}
\end{figure*}

In Table \ref{table:ablation}, we conducted ablation studies on the pipeline and preprocess design of our framework. The results of experiments \#1 and \#5 show that fusing lane line labels as a mask into the Canny image is beneficial, with an increase of 1.68\% in F1@50. The results of experiments \#1, \#2, \#3, and \#4 indicate that using Canny or InstructP2P alone for control, or swapping the order of Canny and InstructP2P in our framework, may lead to suboptimal results. The images listed in Figure \ref{fig:ablation} intuitively demonstrate this point, showing that the labels and semantic information are invisible.

This phenomenon can be likened to the process of painting: the Canny image acts as a geometric sketch, delineating the boundaries for the model and indicating “\emph{where the edges lie},” yet it falls short of pixel-level precision and inevitably introduces noise. InstructP2P supplies the color, informing the model “\emph{what the hues should be},” and is capable of pixel-accurate control. 

When generating night or dusk scenes, sketching the skeleton first and then applying color ensures that the sky tones remain within plausible bounds without importing extraneous noise or spurious features. Reversing the order — coloring before outlining — allows the pigments to overflow the skeletal confines, and the attendant noise seeds unwanted artifacts across the canvas. For adverse-weather conditions such as snow, rain, or fog, we actually desire the emergence of pixel-level stochasticity (snowflakes, rain streaks, misty veils), so the tight pixel-level grip of InstructP2P is deliberately relaxed. 

The rationale for fusing lane-line annotations into the control signal is to preserve label invariance: the lane positions are pre-etched into the geometric skeleton, guaranteeing their immutability throughout the generative process. Positive and negative prompts constrain the model toward the desired generative behavior while discouraging the production of irrelevant content.

It is worth highlighting that, despite the degraded quality of some generated images, the F1@50, F1@75, and mF1 scores remain non-zero. 
This is attributable to the presence of normal-condition samples in the training data, and these samples provide the model with stable prior knowledge.

\subsection{Quality Analysis of Generation}

We also tested the quality of the images generated by our proposed framework and compared it with Stable Diffusion \cite{ro2022sd}, PITI \cite{wang2022piti}, CycleGAN\cite{shao2021cyclegan} and Flux.1 with the results shown in Table \ref{table:quality}. 

\begin{table}[h]
    \centering
    \caption{\textbf{Generative Quality.} The symbol $\uparrow$ indicates that the higher the value of item, the better. The symbol $\downarrow$ is opposite.} 
    \scalebox{0.9}{
    \setlength{\belowrulesep}{0pt}
    \setlength{\tabcolsep}{1.8mm}
    \begin{tabular}{ccccc}
        \Xhline{1.2pt}
        \rowcolor{violet!15} Method & F1@50$\uparrow$ &FID$\downarrow$ &\makecell{CLIP-T\\score}$\uparrow$ & \makecell{Human\\score}\hspace{1.3em}$\uparrow$\\
        \Xhline{1.2pt}
        \rowcolor{gray!10}
        Stable Diffusion & 69.02 & 236.7 & 0.20 & 2.35 $\pm$ 0.75\\
        PITI & 58.50 & 390.5 & 0.16 & 1.1 $\pm$ 0.31\\
        \rowcolor{gray!10}
        CycleGAN & 81.49 & 99.8 & 0.22 & 3.75 $\pm$ 0.55\\
        Flux.1 & 70.04 & 120.3 & 0.21 & 2.5 $\pm$ 0.61\\
        \rowcolor{green!10} \textbf{HG-Lane(Ours)} & \textbf{88.49} &  \textbf{80.1} & \textbf{0.23} & \textbf{4.15 $\pm$ 0.59} \\
        \Xhline{1.2pt}
    \end{tabular}
    }
    \label{table:quality}
\end{table}

We conducted a human survey among 20 participants, inviting them to rate the realism of the generated images on a scale from 1 to 5, where 1 indicates extremely unrealistic and 5 indicates extremely realistic. The resulting Human score is displayed. We used the same generation prompts to test the CLIP-T score. We also tested the F1@50 using CLRNet (ResNet-18) to verify whether the lane line detection performance has changed. The results show that our proposed framework outperforms the others. 

\begin{figure*}[t]
\centering
\includegraphics[width=1\textwidth]{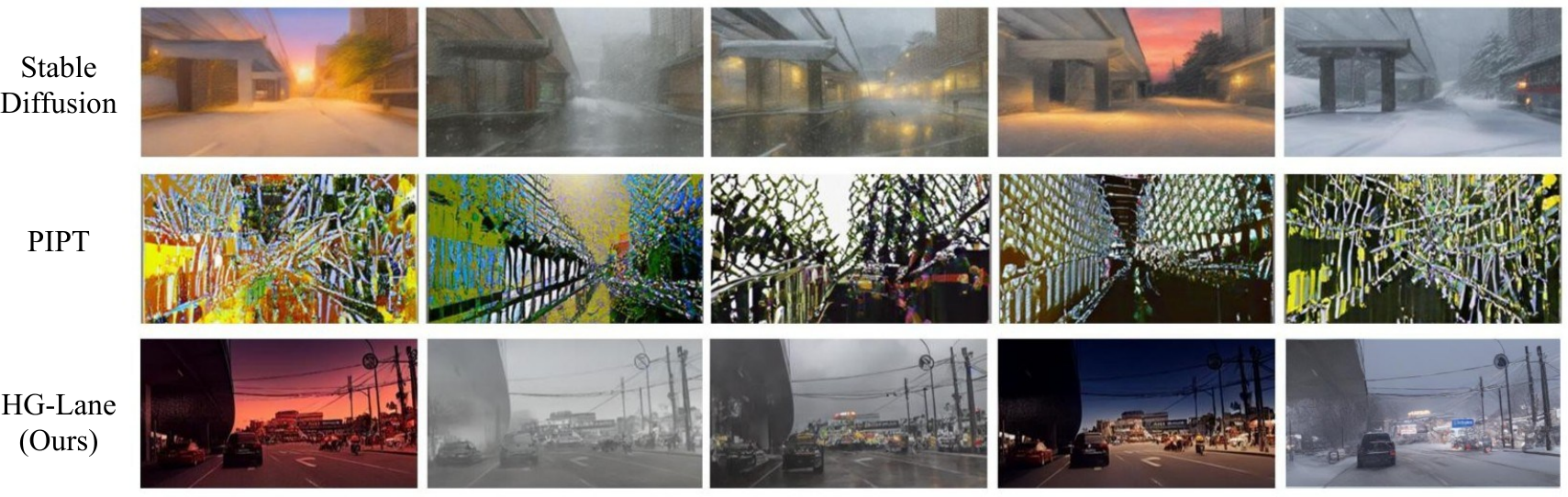}
\caption{\textbf{Quality Analysis of Generation.} Comparison of images generated by different frameworks. The figure illustrates the generation results of different frameworks across five categories (dusk, fog, rain, night, and snow).}
\label{fig:quality}
\end{figure*}

Figure \ref{fig:quality} also demonstrates this point, showing that while Stable Diffusion performs well in preserving lane line labels, it falls short in terms of realism and semantic information retention. On the other hand, PITI leans towards an artistic style, which is not suitable for our task.

\begin{table*}[h]
  \centering
    \caption{\textbf{Performance on CLRNet (ResNet-18) in Real world.} For Normal, Snow, Rain, Fog, Night, and Dusk, the metric used is F1@50.}
  \label{table:realworld}
  \captionsetup{type=table}
  \small
  \setlength{\belowrulesep}{0pt}
  \setlength{\tabcolsep}{2.5mm}
  \begin{tabular}{l|ccccc|ccc}
    \Xhline{1.2pt}
    \rowcolor{violet!15} Train on&Snow&Rain&Fog&Night&Dusk&F1@50&F1@75&mF1\\
    \Xhline{1.2pt}
    CULane&35.41&74.78&50.00&22.47&20.77&53.03&21.62&25.65\\
    \rowcolor{green!10} \textbf{HG-Lane(Ours)}&\textbf{55.65}&\textbf{77.58}&\textbf{65.57}&\textbf{60.02}&\textbf{69.31}&\textbf{68.54}&\textbf{27.48}&\textbf{32.75}\\
    \Xhline{1.2pt}
  \end{tabular}
\end{table*}

\subsection{Performance on Real World}

To evaluate our framework’s generalization ability in the real world, we collected real-world samples, annotated and processed them according to the CULane requirements, and obtained 200 images each for snow, rain, fog, night, and dusk. We then tested them with CLRNet (ResNet-18), and the results are shown in Table \ref{table:realworld}. Compared with the model trained on the same amount of original CULane data, our model achieves substantial improvements on all five categories, with the overall F1@50 increasing by 15.51 \%.

\subsection{Compare with Suppression Module}

To compare the difference between our framework and directly adding weather-suppression modules, we tested our self-collected real-world rain, snow, and fog samples on their respective weather categories using F1@50. 

\begin{table}[h]
    \centering
    \caption{\textbf{Performance on CLRNet (ResNet-18).}} 
    \setlength{\belowrulesep}{0pt}
    \setlength{\tabcolsep}{1.5mm}
    \begin{tabular}{llll}
        \Xhline{1.2pt}
        \rowcolor{violet!15} Category& Method& F1@50& Time(ms/img)\\
        \Xhline{1.2pt}
        DeRain& EfficientDeRain& 74.82& 4.8\\
         & NeRD-Rain& 74.86& 9.2\\
         \rowcolor{green!10} & \textbf{HG-Lane(Ours)}& \textbf{77.58}&\textbf{0}\\
        \Xhline{1.2pt}
        DeSnow& DeSnowNet& 35.21&10.2\\
         & HDCW-Net& 37.21&6.3\\
        \rowcolor{green!10} & \textbf{HG-Lane(Ours)}& \textbf{55.65}&\textbf{0}\\
        \Xhline{1.2pt}
        DeFog& NLD& 51.58&5.1\\
         & GFN& 53.20&7.4\\
        \rowcolor{green!10} &\textbf{HG-Lane(Ours)} & \textbf{65.57}&\textbf{0}\\
        \Xhline{1.2pt}
    \end{tabular}
    \label{table:suppression}
\end{table}
\vspace{-\baselineskip}  

We also measured the overhead time of these suppression modules on an RTX-3090, with results shown in Table \ref{table:suppression}. The modules include EfficientDeRain\cite{qing2021effiderain} and NeRD-Rain\cite{chen2024nerdrain} for deraining, DeSnowNet\cite{liu2018desnownet} and HDCW-Net\cite{chen2021hdcwnet} for desnowing, and NLD\cite{dana2016nld} and GFN\cite{ren2018gfn} for defogging. Our framework demonstrates superior performance across all cases. Besides, our framework works offline, which means it requires no temporal cost during inference.

\section{Conclusion}
In this paper, we present a high-fidelity lane scene generative framework (HG-Lane). It can synthesize realistic adverse-weather and lighting scenes while preserving lane semantics, without the need for re-annotation or fine-tuning. Based on the images generated by our framework, we construct a new benchmark that includes 6 categories (normal, snow, rain, fog, night, dusk), consisting of 30,000 images. The experimental results show that our method effectively improves the performance of detection models. With the SOTA model CLRNet, the overall mF1 on our benchmark increased by 20.87\%. The F1@50 for the overall, normal, snow, rain, fog, night, and dusk categories increased by 19.75\%, 8.63\%, 38.8\%, 14.96\%, 26.84\%, 21.5\%, and 12.04\%. We also provide the F1@50 scores on various baselines, which show excellent performance. Additionally, ablation studies confirm the soundness of our design; comparisons with multiple generative models demonstrate its superiority in preserving label semantics and photorealism; tests on real-world datasets verify its generalization capability; and benchmarking against the suppression module highlights its advantages in tackling the same task.

\noindent\textbf{Future Work:} HG-Lane is designed to preserve lane annotations during generation, which may inevitably introduce some location bias. In the future, we will explore strategies to mitigate this effect while maintaining label consistency.

\section*{Acknowledgments}
This work was supported by the Key Research and Development Promotion Projects in Henan Province (Grant No. 262102210114), the 	National College Student Innovation Training Program (Grant No. 202510357003), and the Open Funding Programs of State Key Laboratory of AI Safety (Grant No. 202507). 


{
    \small
    \bibliographystyle{ieeenat_fullname}
    \bibliography{main}
}

\clearpage
\setcounter{page}{1}
\maketitlesupplementary

\appendix

\section{Visualization in Real-World}
\label{sec:visualization2}

\begin{figure}[h]
\centering
\includegraphics[width=0.5\textwidth]{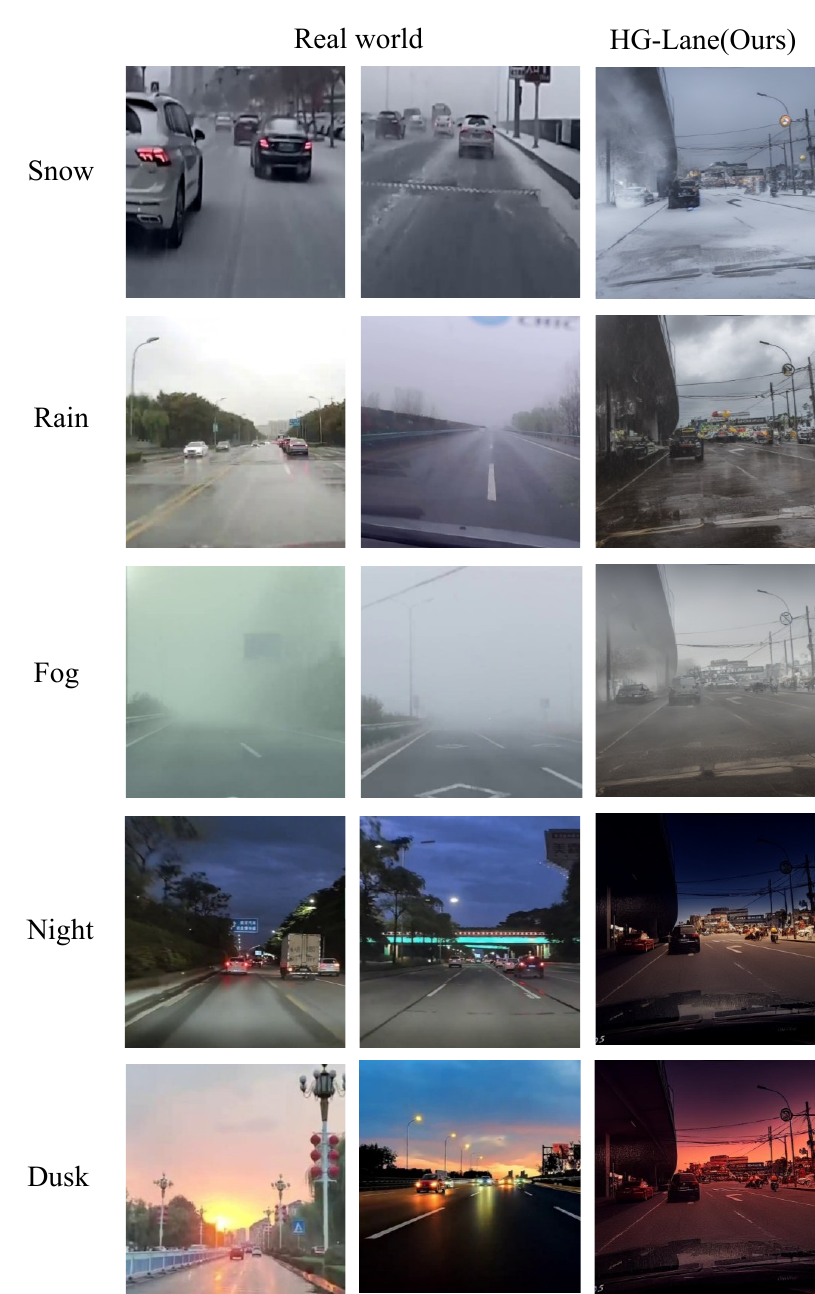}
\caption{\textbf{Comparison with Suppression Module.}}
\label{fig:vis1}
\end{figure}

In Figure \ref{fig:vis1}, a comparison is presented between the samples generated by our framework and real-world samples. We can see that the realism of images generated by HG-Lane is very close to that of real-world images.

\begin{figure}[h]
\centering
\includegraphics[width=0.47\textwidth]{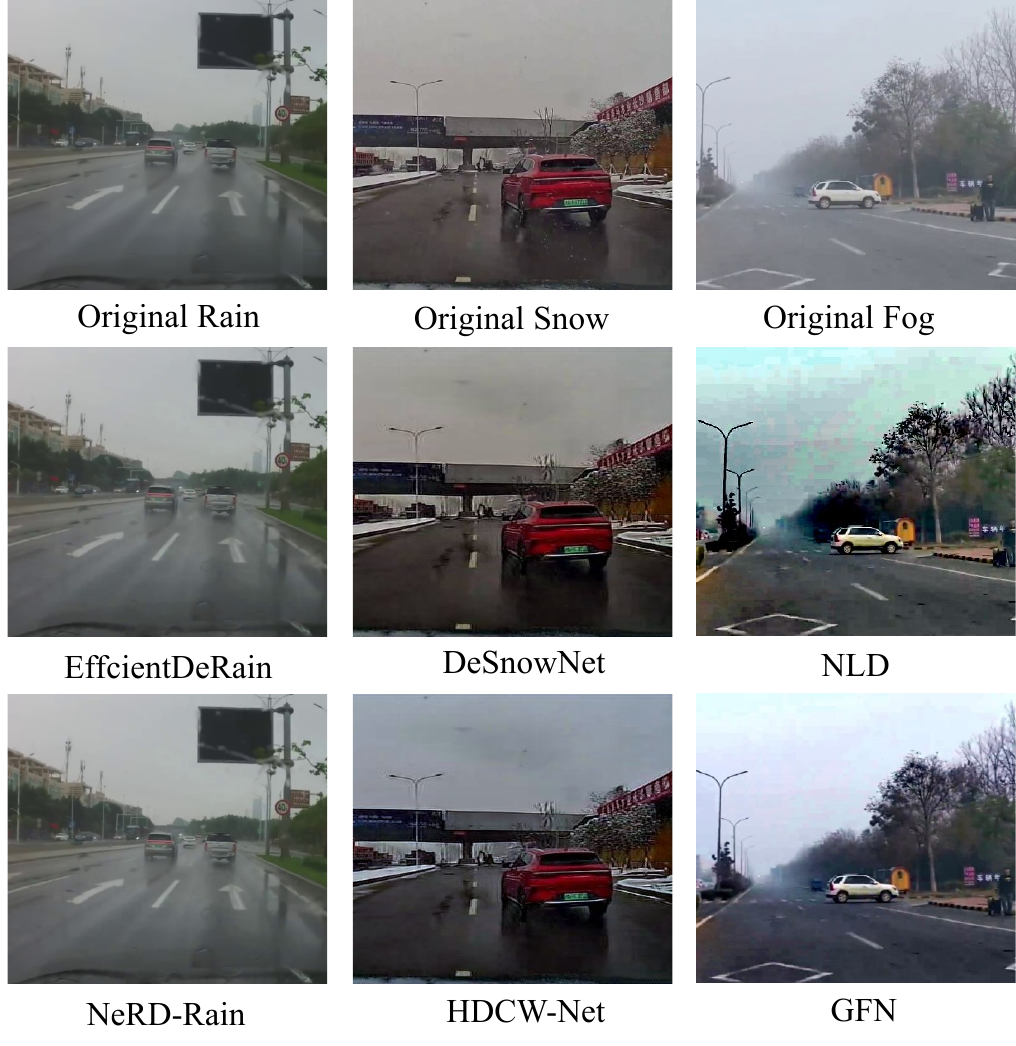}
\caption{\textbf{Comparison with Real-World Samples.}}
\label{fig:vis2}
\end{figure}

\section{Visualization in Suppression Module}
\label{sec:visualization1}
In Figure \ref{fig:vis2}, the samples produced by our framework are compared with those from the suppression module.  We can see that the addition of the suppression module has some effect on removing weather features such as rain streaks, haze, and snowflakes, but it does not fundamentally alter the weather domain. As a result, the lane detection model still exhibits poor generalization performance. Moreover, the suppression module itself incurs additional computational overhead, significantly impacting real-time performance. 

\section{Visualization in Other Dataset}
\label{sec:visualization3}

\begin{figure}[h]
\centering
\includegraphics[width=0.5\textwidth]{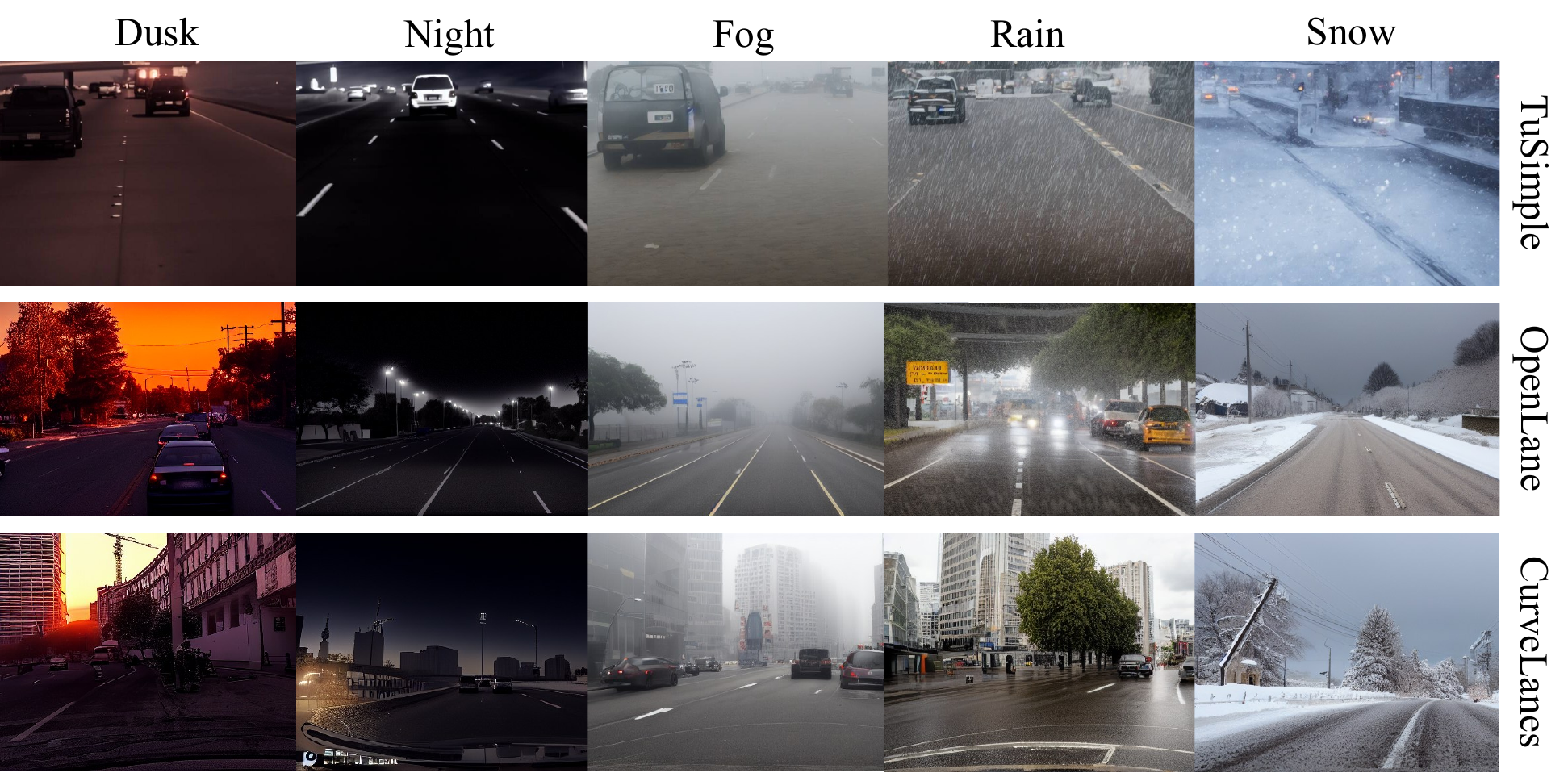}
\caption{\textbf{Examples of other datasets.}}
\label{fig:vis3}
\end{figure}

In Figure \ref{fig:vis3}, we show our framework generalizes well to other mainstream lane detection datasets, such as TuSimple, OpenLane, and CurveLanes, achieving favorable results.

\end{document}